\documentclass[sn-mathphys,Numbered]{sn-jnl}% Math and Physical Sciences Reference Style
\usepackage{graphicx}%
\usepackage{multirow}%
\usepackage{amsmath,amssymb,amsfonts}%
\usepackage{amsthm}%
\usepackage{mathrsfs}%
\usepackage[title]{appendix}%
\usepackage{xcolor}%
\usepackage{textcomp}%
\usepackage{manyfoot}%
\usepackage{booktabs}%
\usepackage{algorithm}%
\usepackage{algorithmicx}%
\usepackage{algpseudocode}%
\usepackage{listings}%
\usepackage{orcidlink}

\begin{document}

\title[VTON-IT: Virtual Try-On using Image Translation]{VTON-IT: Virtual Try-On using Image Translation}

\author*[1]{\fnm{Santosh} \sur{Adhikari }\orcidlink{0000-0003-4994-1090}}  \email{santosh\_adhikari@etu.u-bourgogne.fr }

\author*[2]{\fnm{Bishnu} \sur{Bhusal} \orcidlink{0000-0001-7522-5878}}\email{bhusalb@mail.missouri.edu}

\author[3]{\fnm{ Prashant} \sur{Ghimire} \orcidlink{0000-0003-2891-3285}}\email{ghimireprashant.p@gmail.com}

\author[3]{\fnm{ Anil} \sur{Shrestha} \orcidlink{0000-0001-9782-5874}}\email{anilkumarshrestha.cs@gmail.com}

\affil*[1]{\orgdiv{VIBOT}, \orgname{Université de Bourgogne}, \orgaddress{ \city{Le Creusot}, \country{France}}}

\affil*[2]{\orgdiv{EECS}, \orgname{University of Missouri}, \orgaddress{\city{Columbia}, \postcode{65211}, \state{MO}, \country{USA}}}

\affil[3]{ \orgname{IKebana Solutions LLC}, \orgaddress{\country{Japan}}}


\abstract{Virtual Try-On (trying clothes virtually) is a promising application of the Generative Adversarial Network (GAN). However, it is arduous to transfer the desired clothing item onto the corresponding regions of a human body because of varying body size, pose, and occlusions like hair and overlapped clothes. This paper aims to produce photo-realistic translated images through semantic segmentation and a generative adversarial architecture-based image translation network. We present a novel image-based Virtual Try-On application VTON-IT that takes an RGB image, segments desired body part, and overlays target cloth over the segmented body region. Most state-of-the-art GAN-based Virtual Try-On applications produce unaligned pixelated synthesis images on real-life test images. However, our approach generates high-resolution natural images with detailed textures on such variant images. \footnote{Details of the implementation, algorithms, and codes, are publicly available on Github: https://github.com/shuntos/VITON-IT}
}


\maketitle

\section{Introduction}
\label{Introduction}

Research and Development on Virtual Try-On applications is getting popular as the fashion e-commerce market is rapidly growing. With a virtual try-on application, customers can try the desired cloth virtually before purchasing and sellers can benefit from an increased online marketplace. In addition, this application can reduce the uncertainty of size and appearance that most online shoppers are afraid of. Problems of traditional 3D-based virtual try-on are computational complexity, tedious hardware-dependent data acquisition process, and less user-friendly \cite{hauswiesner2011free}. Image-based 2D virtual try-on applications, if integrated into existing e-commerce or digital marketplace, will be more scalable and memory efficient compared to the 3D approach.

In the recent advancements in GANs, image-to-image translation in the conditional setting has become possible \cite{isola2017image}. Improved discriminator and generation architectures have enabled cross-domain high-resolution image translation \cite{high_resolution_image_syn}, allowing for the transformation of styles and textures from one domain to another.

In this paper, we primarily address the challenges of 2D virtual try-on applications by leveraging state-of-the-art deep learning networks for semantic segmentation and robust image translation networks for translating input images into the target domain. Previous works like VVT \cite{dong2019fw} encountered issues with semantic segmentation due to plain backgrounds in training datasets. To address this, we trained a UNet-like semantic segmentation architecture on diverse images manually selected from the FGV6 dataset \cite{iMateria0:online}. For image translation tasks, a residual mapping generator and multi-scale discriminator are employed, taking a semantic mask from the segmentation network and translating it into a wrapped RGB cloth with fine details. Previous methods only worked on images with a single person \cite{liu2021arbitrary}. Thus, to make it applicable for multi-person cases, we utilized a pre-trained human detection Yolov5 model \cite{glenn_jocher_2021_5563715} trained on the COCO dataset \cite{lin2014microsoft} to generate bounding boxes for each human body and crop the overlaying cloth accordingly.

The VTON-IT architecture offers pose, background, and occlusion invariant applications for the online fashion industry with a wide range of applications. Rigorous testing and experiments have demonstrated that our approach generates more visually promising overlayed images compared to existing methods.

\section{Related Works}
\label{sec:related_works}

Several image-based virtual try-on approaches have been tried in the prior research, those relevant in our study are discussed here.

\subsection{VITON}
Han et al. presented an image-based Virtual Try-On Network (VITON) \cite{han2018viton}: a coarse-to-fine framework that seamlessly transferred a target clothing item in a product image to the corresponding region of a clothed person in a 2D image. The warped target clothing to match the pose of the clothed person was generated using a thin-plate spline (TPS) transformation which is ultimately fused with the person's image.

\subsection{CP-VTON}

CPVTON \cite{wang2018toward} adopts a structure similar to VITON, but it utilizes a neural network to learn the spatial transformation parameters of the TPS transformation within its Geometric Matching Module (GMM). Consequently, the GMM generates a warped cloth image, and a try-on module fuses the warped cloth image with the target image of the person, preserving the precise features of the clothes.

\subsection{CP-VTON+}
CP-VTON+ \cite{CP_VTON_plus} proposed a framework that preserves both cloth shape and texture through a two-stage architecture. The first stage is the Clothing Warping Stage, which transfers the texture of the clothing from the clothing image to the target person image. The later stage is the Blending Stage, which introduces a refinement module to further improve the quality of the generated try-on image. The framework consists of four major components: a body parsing network, a spatial transformer network, a shape transfer network, and a texture transfer network.

\subsection{VTNFP}
VTNFP \cite{yu2019vtnfp} adopts a three-stage design strategy. Initially, it generates warped clothing, followed by generating a body segmentation map of the person wearing the target clothing. Finally, it employs a try-on synthesis module to fuse all information for the final image synthesis. This method effectively preserves both the target cloth and human body parts, ensuring that clothes requiring no replacement remain intact.

\subsection{Virtual Try-on auxiliary human segmentation}
Virtual Try-On using auxiliary human segmentation \cite{ayush2019powering} builds upon the existing CP-VTON framework by incorporating additional enhancements. Leveraging human semantic segmentation predictions as an auxiliary task significantly enhances virtual try-on performance. The proposed architecture introduces a branched design to concurrently predict the try-on outcome and the expected segmentation mask of the generated try-on output, with the target model now adorned in the in-shop cloth.

\section{Proposed Approach}
\label{Proposed Approach} 

The network architecture proposed in this study comprises three key components. The first stage involves implementing a human body parsing network, followed by a body region segmentation network. Finally, an image translation network is employed to facilitate the wrapping of input clothing over the target body.

% \FloatBarrier
\begin{figure}[ht]
  \centering
  % include first image
  \includegraphics[width=1\linewidth]{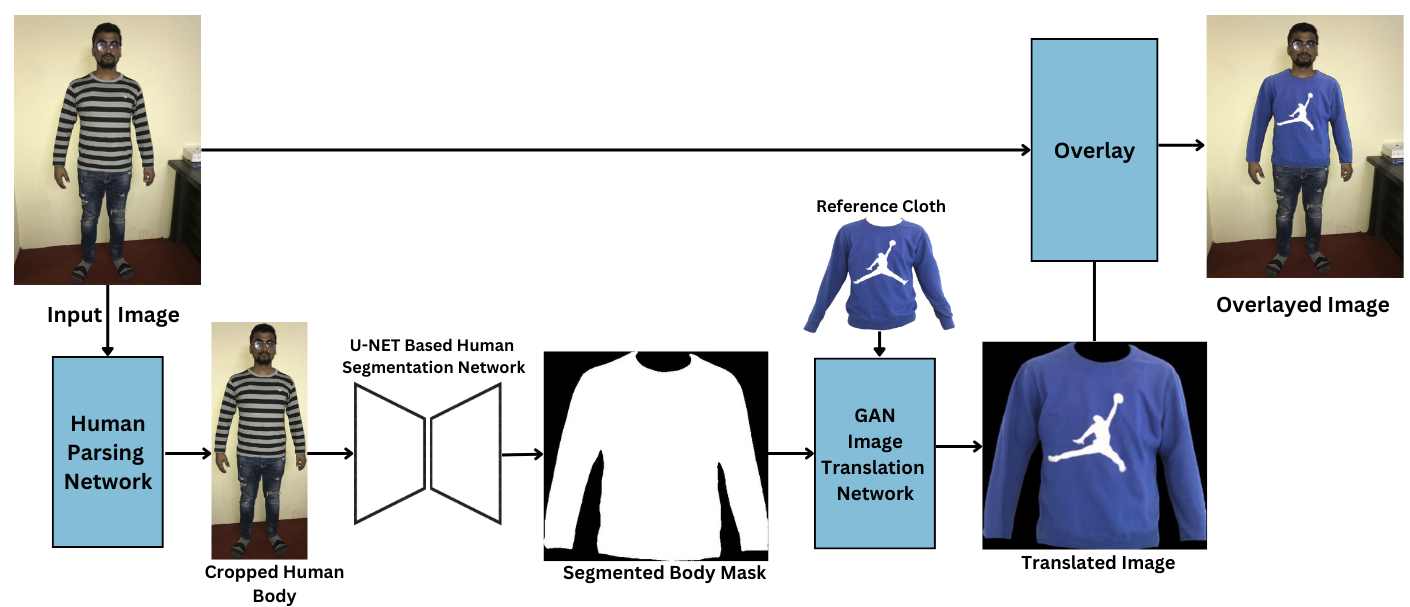 } 
\caption{Proposed VITON-IT overview. First, the human body is detected and cropped. Then, the desired body region is segmented through $U^2$-Net architecture and the segmented mask is fed to the image translation network to generate wrapped cloth. Finally, the wrapped cloth is overlayed over the input image.}
\label{fig:proposed_approach}
\end{figure}%

\subsection{Human Parsing Network}
To parse the body part for translation, the input image is given as input to the Yolov5 pre-trained model trained on the Microsoft COCO dataset  \cite{lin2014microsoft} for object detection tasks. This model outperforms existing object detection models in terms of latency and memory consumption. The model gives bounding boxes of humans in image and probability score. This allows the user to perform image translation on multiple objects. Pytorch implementation of the Yolov5-large model is used for inference. Input image of size $(640 \times 640)$ is fed into the network with confidence threshold $=0.25$, Non-Maximum IOU threshold $=0.45$, and $max\_detection=10$ is used.

\subsection{Human body Segmentation}

The $U^2$-Net architecture has been implemented to generate masks of desired body parts for image translation tasks. While existing backbones like AlexNet \cite{krizhevsky2017imagenet}, VGG \cite{simonyan2014very}, ResNet \cite{he2016deep}, and DenseNet \cite{huang2017densely} are utilized for semantic segmentation tasks, their feature maps ~\cite{inbook} have lower resolutions. For instance, ResNet reduces the size of feature maps to one-fourth of the input size. However, feature map resolution is crucial for salient object detection (SOD), where the objective is to segment the most visually appealing object in an image. Additionally, these backbones often have complex architectures due to the inclusion of additional feature extractor modules for extracting multi-level saliency features.

The Nested UNet architecture, $U^2$-Net, aims to delve deeper while preserving feature map resolution. It can be trained from scratch and maintains the resolution of feature maps with the help of residual U-blocks (RSU). A top-level UNet-like architecture is followed by RSU, which can extract intra-stage multi-scale features.

\textbf{Architecture}: In salient object detection tasks, both local and global contextual information are crucial. Existing feature extractor backbones tend to have small receptive fields, using small convolution filters of size $(1 \times 1)$ or $(3 \times 3)$ for computational efficiency and storage space considerations. However, such small receptive fields may struggle to capture global information. Enlarging the receptive field can be achieved through dilated convolution operations. Nonetheless, multiple dilated convolution operations at the original resolution require significant computational and memory resources. To address this issue, RSU is employed, consisting of three main components.

\begin{enumerate}
    \item Input Convolution layer: Transforms input feature $map( H, W, Cin)$  to intermediate feature map $F1(x)$ with $cout$. Local features are extracted from this plain convolution layer. 
    \item Input feature map $F1(x)$ is given as input to UNet like the architecture of height $L$, a higher value of $L$ means deeper the network with a large number of pooling layers and ranges of receptive fields and numbers of local and global features. During downsampling multi-scale features are extracted and higher-resolution feature maps are encoded through progressive upsampling, concatenation, and convolution.  
    \item Fusion of local and global features is done through RSU.
	$H(x) =(F(x)+U(F(x))$
Where $F(x) =$ Intermediate feature map, $U(F(x)) =$ Multi-Scale contextual information, and $H(x)=$ Desired mapping of input features.

\end{enumerate}

$U^2$-Net consists of two-level nested U structures. The first level consists of 11 well-configured RSU stages capable of extracting intra-stage multi-scale and inter-stage multi-level features. Architecture has main 3 parts: a six-stage encoder, a five-stage decoder, and a saliency map fusion module with a decoder stage and the last encoder stage. 
In the first four encoder stages, RSU is used but in the 5th and 6th stage resolution of the feature map is relatively low, so further downsampling might lead to loss of contextual information. Thus in the 5th and 6th encoder stages dilated RSU (replaced upsampling and pooling with dilated convolution) is used to preserve the resolution of feature maps. Feature map resolution in the 4th to 6th stages is the same. 
In decoder stages, dilated RSU is used when each stage takes concatenated upsampled feature maps from the previous stage. 
The saliency map fuser inputs saliency probability maps from five decoders and the last stage (6th) encoder where each map is generated through $(3 \times 3)$ convolution and sigmoid function. These probability maps are concatenated and passed through $(1 \times 1)$ convolution and sigmoid function to generate the final saliency probability map \(S_{fused}\).

\subsection{Image Translation}
Masks generated by the human body segmentation network are passed through an image translation network (pix2pix). This network produces high-resolution photo-realistic synthesis RGB images from semantic label maps by leveraging a Generative Adversarial Network (GAN) in a conditional setting.  Such visually appealing images are produced under adversarial training instead of any loss functions.
\\ \\
\textbf{Architecture}: This GAN framework consists of generator $G$ and discriminator $D$ for image-image translation tasks. 
The task of generator $G$ is to generate an image(RGB) of cloth given a binary semantic map generated by the human body segmentation network. Whereas, discriminator $D$ tries to classify whether the generated image is real or synthesized. The dataset consists of a pair of images ($S_i$ and $X_i$) where $S_i$ is a mask and $X_i$ is the corresponding real image. This architecture works in supervised configuration to model the conditional distribution of real images with given binary masks with a min-max game. Where generator $G$ and discriminator $D$ try to win against each other. From \cite{gan_original_paper}, we have,

\begin{equation}
\min _{G} \max _{D} V(D, G)=\mathbb{E}_{\boldsymbol{x} \sim p_{\text {data }}(\boldsymbol{x})}[\log D(\boldsymbol{x})]+\mathbb{E}_{\boldsymbol{z} \sim p_{\boldsymbol{z}}(\boldsymbol{z})}[\log (1-D(G(\boldsymbol{z})))]
\label{eq:min_max_G}
\end{equation}

This architecture uses UNet as a generator and a patch-based network as a discriminator. The generator takes 3 channel mask whereas a concatenated channel-wise semantic label map and the corresponding image are fed to Discriminator. The main components of this architecture are a coarse-to-fine generator, a multi-scale discriminator, and an optimized adversarial objective function. 
The generator consists of a global generator network $G1$ and a local enhancer network. The local generator outputs an image of resolution $4$ times larger than the original, or in other words ( 2 times of width and height)  greater than the previous one. To increase the resolution of the synthesis image additional local generators can be added. For instance, the output resolution of $\{G1, G2)$ is $1024\times2048$ whereas the output resolution of $\{G1, G2, G3\}$ is $2048\times4096$. 
First global generator $G1$ is trained and then the local generator  $G2$ is trained and so on, in the order of their resolution. Global generator $G1$ is trained on low-resolution images then another residual network $G2$ is added to $G1$ and a joint network is trained on higher-resolution images. Element wise sum of the feature map of $G2$ and the last feature map of $G1$ is fed into the next $G2$.

To differentiate real image and synthesis image, the discriminator must have a greater receptive field to capture global contextual information and similarly be able to extract lower-level local features. Discriminator architecture consists of $3$ different identical discriminators each working on different resolutions. The resolution of the real and synthesized image is downscaled by the factor of $2$ to create an image pyramid of $3$ scale.  By this approach, a discriminator with the finest resolution helps the generator to produce an image with fine details.

\begin{equation}
   \min _{G} \max _{D_{1} D_{2} D_{3}} \sum_{k=1,2,3} \mathcal{L} _{G A N}\left(G, D_{k}\right)
\end{equation}

Where $D_1 D_2, D_3$ are $3$ scale discriminators \cite{high_resolution_image_syn}. As this architecture uses a multi-scale discriminator it extracts features from its multiple layers thus the generator has to produce a natural image at multiple scales. By adding such feature loss in the discriminator this helps to stabilize training loss. 

Feature match loss is given by \cite{high_resolution_image_syn},

\begin{equation}
\mathcal{L}_{GM}(G, D_k) = \mathbb{E}_{(s, x)} \sum_{i =1}^T \frac{1}{N_i}[||D_k^{(i)}(\boldsymbol{s}, \boldsymbol{x}) - D_k^{(i)} (\boldsymbol{s}, G(\boldsymbol{s})) ||_{1} ]
\end{equation}

Where $D_k$, $D_k^{(i)}$ are feature extractor layers of the discriminator, $T$ represents the total number of layers and $N$ represents the number of elements per layer. 

By adding feature match loss and GAN loss objective function becomes \cite{high_resolution_image_syn},

\begin{equation}
    min_G \left( \left( max_{D_1, D_2, D_3} \sum_{k=1,2,3} \mathcal{L} _{G A N}\left(G, D_{k}\right) \right)  + \lambda \sum_{k = 1, 2, 3} \mathcal{L}_{FM}(G, D_k) \right)
\end{equation}

This loss function works well for translating mask to cloth image with higher resolution and detailed texture.

% \FloatBarrier
\begin{figure}[ht]
  \centering
  % include first image
  \includegraphics[width=1\linewidth]{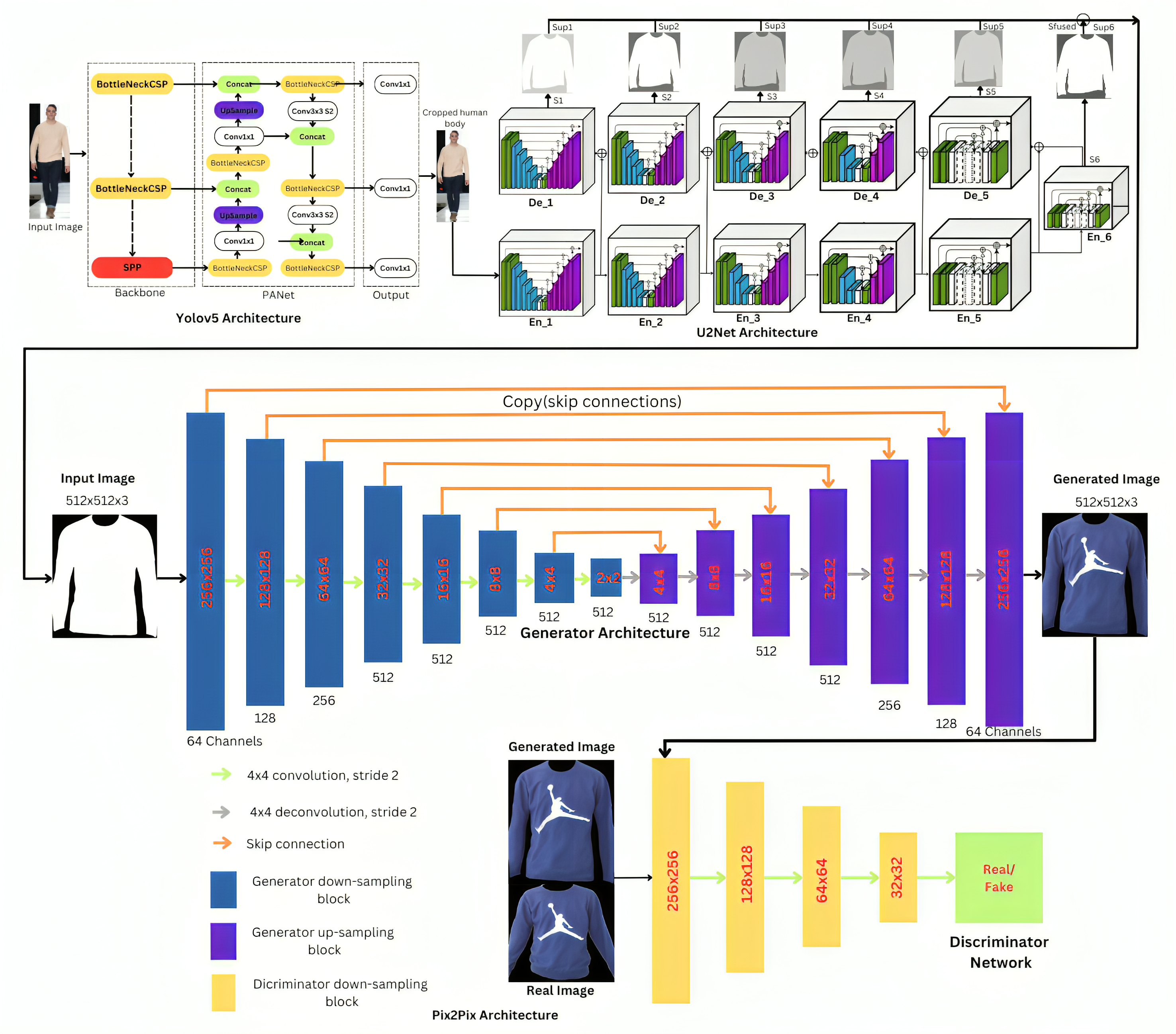} 
\caption{Virtual Try-On Architecture: An input image is first fed to the YOLOv5 object detection model to detect the human body, which is then cropped. The cropped image is then passed through the $U^2$-Net segmentation model to generate a body region mask. Finally, the mask is fed into the Pix2Pix generator, which synthesizes RGB-based clothing onto the masked body region, resulting in a virtual try-on of the clothing. The output image shows the synthesized clothing on the original human body image.}
\label{fig:overall_architecture}
\end{figure}%

\section{Implementation Details}

\subsection{Training Human Body Segmentation network}

For the training dataset, $6000$ good quality images are selected manually from FGVC6 \cite{iMateria0:online} dataset and labeled using Labelme tool \cite{russell2008labelme}  to generate desired body masks. The average resolution of training images is $(630 \times 1554)$. 
The model was trained through transfer learning using a pre-trained model trained on the COCO dataset for general human body segmentation task with input Image size $320 \times 320$ with random flip and crop.  
Pytorch library is used for training and inference. Adam optimizer \cite{adam_optimization} is used to train our network and its hyperparameters are set to default (initial learning rate $lr=1e-3$, $betas=(0.9, 0.999)$, $eps=1e-8$, $weight\_decay=0$). Initially, the loss is set to $1$. The total number of iterations was $400000$ with a training loss of $0.109575$.

The model was trained on a custom dataset with unique ground truth. So performance was evaluated on a custom test dataset and we achieved $MaxFB = 0.865$, Mean average Error $(MAE)= 0.081$, $FBw=0.801$, and $SM= 0.854$.

\subsection{Training Image Translation Network}

The dataset was prepared manually by creating a pair of real images and a corresponding mask. Images are labeled using the Labelme tool to generate a semantic label and apply a series of geometrical augmentation algorithms. We prepared training pair images through data augmentation. As deep neural networks involve millions of parameters, incorporating more training data that is relevant to the domain can effectively mitigate the problem of overfitting. According to Zhao et. al \cite{image_aug_for_gan}, GAN performance is improved more by augmentations that cause spatial changes than by augmentations that just cause visual changes. Therefore, we have used several geometric augmentation techniques to augment the cloth image before training. These are namely Perspective Transform, Piecewise Affine Transform,  Elastic Transformation, Shearing, and Scaling. We augmented both the image and its corresponding mask using the Imgaug library \cite{imgaug}. The image translation model was trained on a 3-channel input image of size $(512\times512)$ with a batch size of $4$, without input label and instance map. The generator generates output images with the same number of image channels and shapes. Final loss consists of three components: GAN loss, discriminator-based feature matching loss, and VGG perceptual loss. After training up to $100$ epoch we got a GAN loss of $0.83$, a GAN Feature loss of $2.123$, and a VGG perceptual loss of $1.81$.

% \FloatBarrier
\begin{figure}[ht]
  \centering
  % include first image
  \includegraphics[width=1\hsize, height=5cm]{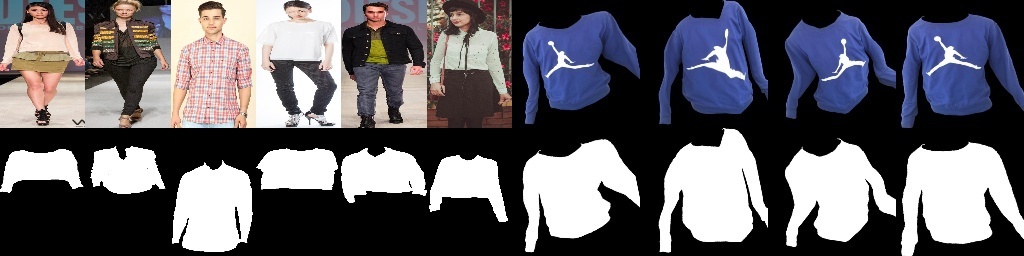 } 
\caption{Example training image for human segmentation and image translation network}
\label{fig:pix2pix_train_ex}
\end{figure}%

% \FloatBarrier
% \begin{figure}[ht]
%   \centering
%   % include first image
%   \includegraphics[width=1\linewidth]{./pix2pix_training_pair.jpg } 
% \caption{Image Translation Training Pair Images}
% \label{fig:pix2pix_train_ex}
% \end{figure}%

% \FloatBarrier
% \begin{figure}[ht]
%   \centering
%   % include first image
%   \includegraphics[width=1\linewidth]{./u2net_train_image.jpg } 
% \caption{Human Body Segmentation Training Images(Top) and Corresponding Masks(Bottom)
% }
% \label{fig:u2net_train_ex}
% \end{figure}%
% \FloatBarrier

\subsection{Training Setup}
Both human segmentation and image translation network were trained on a Ubuntu 16.04 with 2 Nvidia GPU: GA102 [GeForce RTX 3090], 32 GB RAM, and  CPU: 20 core Intel(R) Core(TM) i9-10900K CPU @ 3.70GHz.

\begin{center}
\begin{figure}[!ht]
    \centering

    %% Added bggroup for spacing between cells
    \bgroup
    % specify spaces for rows
    \def\arraystretch{1} 
    % specify spaces for cols
    \setlength\tabcolsep{2pt}
    
    \begin{tabular}{ c c c c c c } 
    
    {} & {} & 
    \multicolumn{2}{c}{CP-VTON+} & 
    \multicolumn{2}{c}{VTON-IT(proposed)}  
     \\
     Reference & 
     Target &
     Wrapped &
     Final &
     Wrapped &
     Final 
     \\
     Image & 
     Clothes &
     Clothes &
     Result &
     Clothes &
     Result 
     \\
     \includegraphics[width=0.15\hsize]{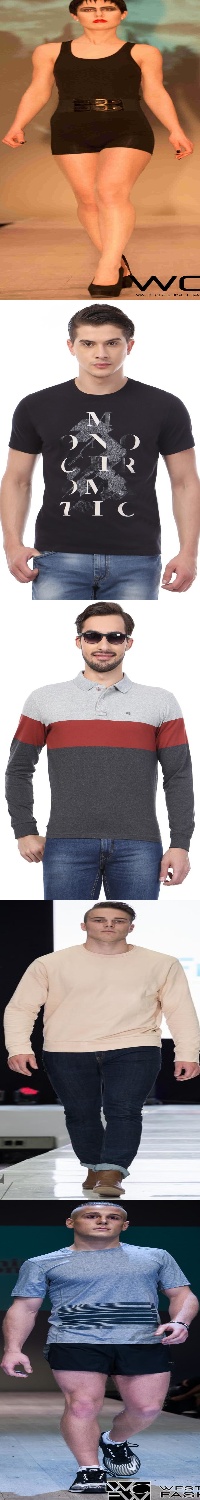} & \includegraphics[width=0.15\hsize]{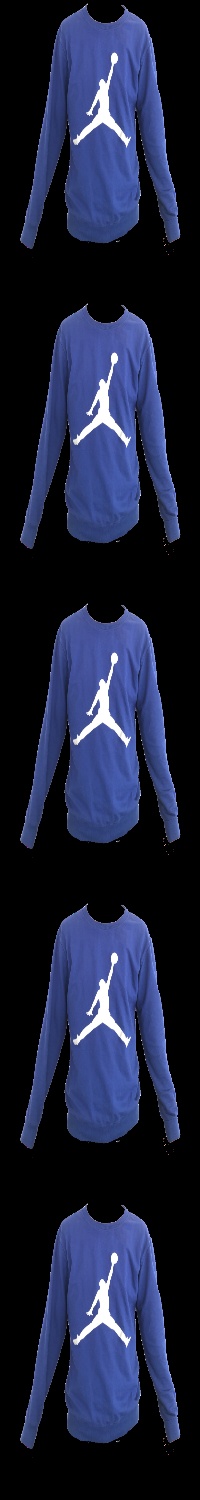} & \includegraphics[width=0.15\hsize]{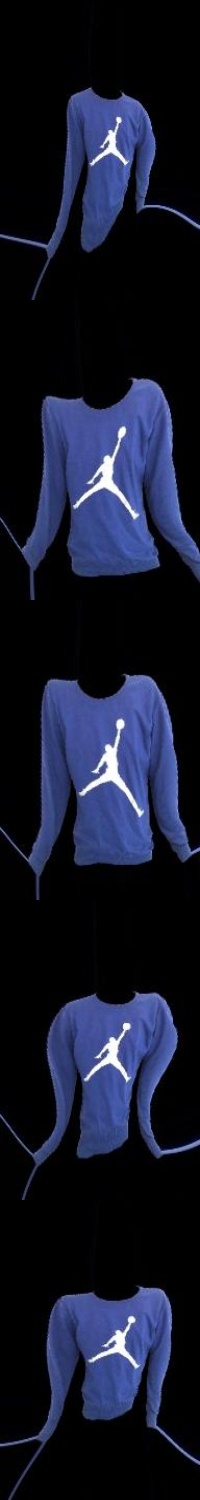} & \includegraphics[width=0.15\hsize]{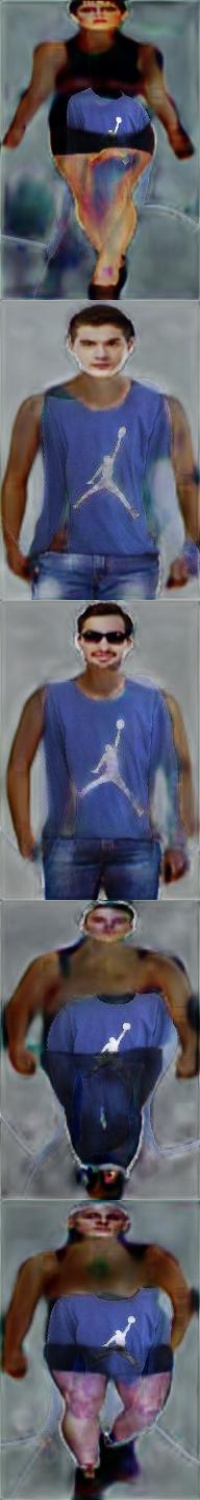} & \includegraphics[width=0.15\hsize]{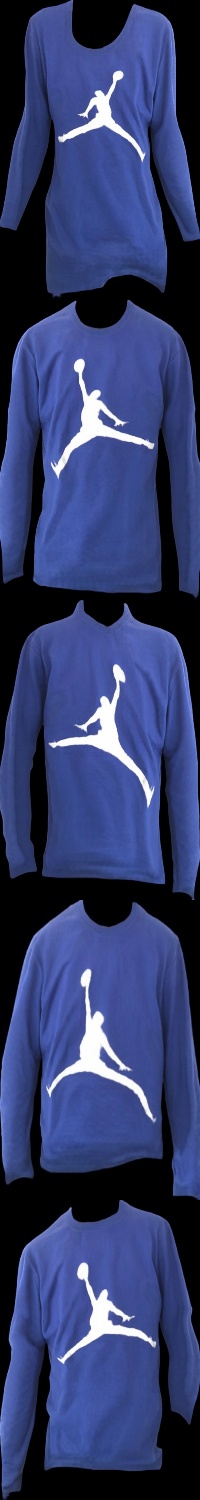} & \includegraphics[width=0.15\hsize]{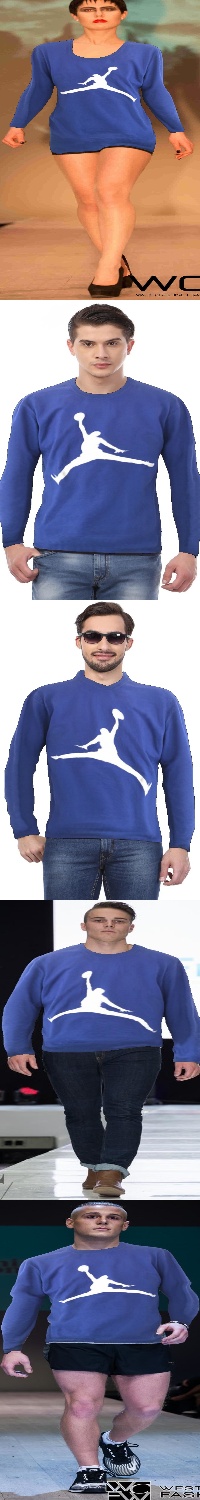}   \\

    \end{tabular}
    \egroup
    
    \caption{Visualized comparison with CP-VTON+}
    \label{fig:Qualitative comparisons on CP-VITON+ (3-4th rows) and VITON-IT(ours) (5-6th rows)}
\end{figure}
\end{center}

% \FloatBarrier
\begin{figure}[!ht]
  \centering
  \includegraphics[width=1\linewidth, height=3cm]{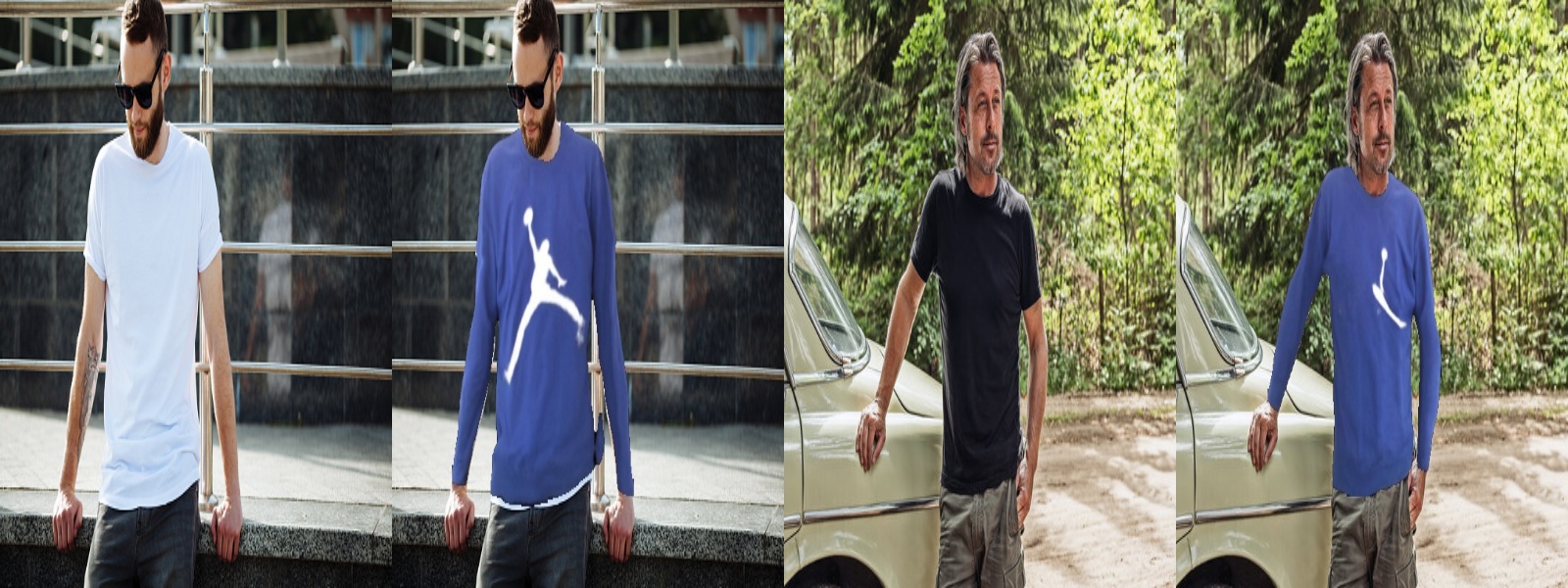}  
\caption{Result on outdoor images}
\label{fig:Qualitative comparisons on CP-VITON+ (3-4th rows) and VITON-IT(ours) (5-6th rows) outdoors}
\end{figure}%

\section{Experimental Results}
\label{results}

\subsection{Qualitative Results}

For evaluating the performance of VTON-IT through visual observation, we compared the final overlayed images with the output of CP-VTON+  \cite{CP_VTON_plus}. Figure~\ref{fig:Qualitative comparisons on CP-VITON+ (3-4th rows) and VITON-IT(ours) (5-6th rows)} shows that the proposed virtual try-on application produces more realistic and convincing results in terms of texture transfer quality and pose preservation. Most of the existing virtual try on produce low-resolution output images. CP-VTON+ generates an output image with a fixed shape $(192 \times  256)$ but our proposed approach works on high-resolution images. Through a high-resolution image translation network wrapped cloth of shape $(512 \times 512)$ is generated. While experimenting with high-resolution input of shape $(2448 \times 3264)$ we got a perfectly aligned natural-looking overlayed image with the same shape as input.

\subsubsection{Result on Outdoor Images}
Even though most of the images used for training body segmentation and translation are captured indoors with proper lighting conditions and predictable poses, both segmentation and translation models produce promising results on outdoor images with noisy backgrounds, unusual poses, and different lighting conditions. Figure~\ref{fig:Qualitative comparisons on CP-VITON+ (3-4th rows) and VITON-IT(ours) (5-6th rows) outdoors} shows the results of the inference performed on outdoor images. However, the image on the right side has some artifacts in the unusual pose.

\subsection{Quantitative Results}

We adopted Structural Similarity Index(SSIM) \cite{wang2004image}, Multi-Scale Structural Similarity (MS-SSIM), Fréchet Inception Distance (FID) \cite{heusel2017gans}, and Kernel Inspection Distance (KID) \cite{binkowski2018demystifying} scores to measure the similarity between ground truths and synthesized images. The ground truths were made by manually wrapping clothes over the models’ images using various imaging tools and the synthesized images were the output of the model used. The results are shown in Table ~\ref{eval-table}.

\subsection{User Study}
Although SSIM, MS-SSIM, FID, and KID can be used to determine the quality of image synthesis, it cannot reflect the overall realism and visual quality as assessed by human evaluation. Thus we performed a user study with 60 volunteers. To evaluate realism, volunteers were provided with two sets of images: ground truth images (manually wrapped clothes on the human models) and the outputs generated by our model. They were asked to score based on how real the clothes looked on the person and how well the texture of the clothing was preserved. Then, they were asked to independently rate the photo realism of only our output images. The result shows that our result was 70\% similar to the ground truth and 60\% photo-realistic.

\begin{table}[ht]
\caption{Quantitative evaluation of CP-VTON+ and VTON-IT in terms of SSIM, MS-SSIM, FID, and KID scores. }\label{eval-table}
\centering
\begin{tabular}{|l|l|l|l|l|}
\hline
Methods &  SSIM & MS-SSIM & FID & KID\\
\hline
CPVTON+ & 0.83 & 0.60 & 393 & 0.077 \\
VTON-IT & \textbf{0.93} & \textbf{0.87} & \textbf{50} & \textbf{0.019}\\
\hline
\end{tabular}
\end{table}

\section{Discussion}
% Experiments on this paper have shown the effectiveness of this approach in both male and female bodies, different lighting conditions, occlusions (hand, hair, etc) and variant poses. The first stage human detection network plays an important role to improve overall performance of application by reducing input image size for later stages and  eliminating unnecessary input regions by cropping the human body. To generate a geometrically-correct segmentation map we need to train a human body segmentation network with  ground truth images having detailed wrist and neck regions. Problem with public dataset like LVIS \cite{gupta2019lvis}, MS COCO \cite{lin2014microsoft} and, Pascal Person Part dataset \cite{chen2014detect} is improper ground truth, hence we tried different pre-trained models like CDCL \cite{lin2020cross} , GRAPHONY \cite{Gong2019Graphonomy}, U2NET \cite{Qin_2020_PR} to generate ground truth but none of these generate precise body mask. So, we manually picked 6000 good images of both men and women from the FGC6 dataset and labeled them manually. We trained a Generative image translation model on conditional settings with a pair of images as input. To generate a pair of images we perform geometrical augmentation on both the semantic mask produced by the segmentation network and its corresponding real RGB image. 

The effectiveness of this approach has been demonstrated through experiments conducted on both male and female bodies, under various lighting conditions, and with occlusions such as hands and hair, as well as varying poses. The initial stage of the human detection network plays a crucial role in enhancing the overall performance of the application by reducing the input image size for subsequent stages and eliminating unnecessary input regions through cropping the human body.

To create an accurate segmentation map adhering to geometric principles, it is essential to train a human body segmentation network using ground truth images containing detailed wrist and neck regions. However, existing public datasets like LVIS \cite{gupta2019lvis}, MS COCO \cite{lin2014microsoft}, and Pascal Person Part dataset \cite{chen2014detect} suffer from improper ground truth annotations. Therefore, we attempted to use various pre-trained models like CDCL \cite{lin2020cross}, GRAPHONY \cite{Gong2019Graphonomy}, and U2NET \cite{Qin_2020_PR} to generate ground truth annotations, but none of these produced precise body masks. Consequently, we manually curated 6000 high-quality images of both men and women from the FGC6 dataset and labeled them manually.

We trained a generative image translation model in conditional settings with a pair of images as input. To generate a pair of images, we utilized geometric augmentation techniques on both the semantic mask generated by the segmentation network and the corresponding real RGB image.

% To generate a pair of images we perform geometrical augmentation on both the semantic mask produced by the segmentation network and its corresponding real RGB image. 

\section{Conclusion and Future Works} \label{analysis}

The paper introduces an innovative approach named VTON-IT (Virtual Try-On using Image Translation) that facilitates the transfer of desired clothing onto a person's image while accommodating variations in body size, pose, and lighting conditions. This method surpasses existing approaches, as evidenced by both quantitative and qualitative results showcasing the generation of natural-looking synthesized images. The proposed architecture comprises three components: human detection, body part segmentation, and an image translation network. While this paper focuses on training the image translation network to generate synthesized images within the same domain (specifically, sweatshirts), it can be adapted for cross-domain synthesis by incorporating control parameters as label features. Future work could extend this approach to different types of clothing such as trousers, shorts, shoes, and beyond.

\textbf{Acknowledgements}: The authors extend their gratitude to IKebana Solutions LLC for their constant support throughout this research project.


\bibliography{main}% common bib file

\end{document}